\documentclass{article}
\usepackage{iclr2027_conference,times}
\usepackage[T1]{fontenc}
\usepackage[utf8]{inputenc}
\usepackage{amsmath,amssymb,amsthm,booktabs,array,multirow}
\usepackage{graphicx,xcolor,tikz,algorithm,algpseudocode}
\usepackage{microtype,url,xurl,hyperref,flafter,placeins,capt-of}
\usetikzlibrary{arrows.meta,positioning,calc,fit}
\definecolor{flowblue}{HTML}{2563A6}
\definecolor{flowgreen}{HTML}{147D74}
\definecolor{floworange}{HTML}{BD7224}
\hypersetup{colorlinks=true,citecolor=flowblue,linkcolor=flowblue,urlcolor=flowblue,pdfauthor={Songhe Wang, Lifu Wei, Shuolin Xu, Charles A. Kamhoua, David Miller},pdftitle={TripleFlow: Training-Free Video Object Removal by Bridging Residual Editing and Native Generation}}
\newcommand{\method}{\textsc{TripleFlow}}

\title{TripleFlow: Training-Free Video\\Object Removal by Bridging Residual\\Editing and Native Generation}
\iclrfinalcopy
\author{%
\begin{minipage}[t]{\dimexpr\textwidth-2\tabcolsep\relax}
\raggedright
\normalsize\bfseries
Songhe Wang\textsuperscript{1,*}\quad
Lifu Wei\textsuperscript{2,*}\quad
Shuolin Xu\textsuperscript{3}\\
Charles A. Kamhoua\textsuperscript{4}\quad
David Miller\textsuperscript{5}\\[0.45em]
\normalfont\small
\textsuperscript{1}CSE Department, Penn State University\\
\textsuperscript{2}Department of Computer Science, The University of British Columbia, Canada\\
\textsuperscript{3}Department of Computing, Bournemouth University, United Kingdom\\
\textsuperscript{4}DEVCOM Army Research Laboratory, Network Security Branch, Adelphi, MD\\
\textsuperscript{5}EE Department, Penn State University\\
\textsuperscript{*}Equal contribution.\\[0.4em]
\textbf{Project page:} \url{https://songhe1998.github.io/TripleFlow/}
\end{minipage}%
}
\begin{document}
\maketitle
% Neutral preprint header; no conference acceptance or review-status claim.
\lhead{TripleFlow}
\rhead{}
\fancyfoot[L]{}
\fancyfoot[R]{}
\begin{center}
\begin{minipage}{0.85\linewidth}
\input{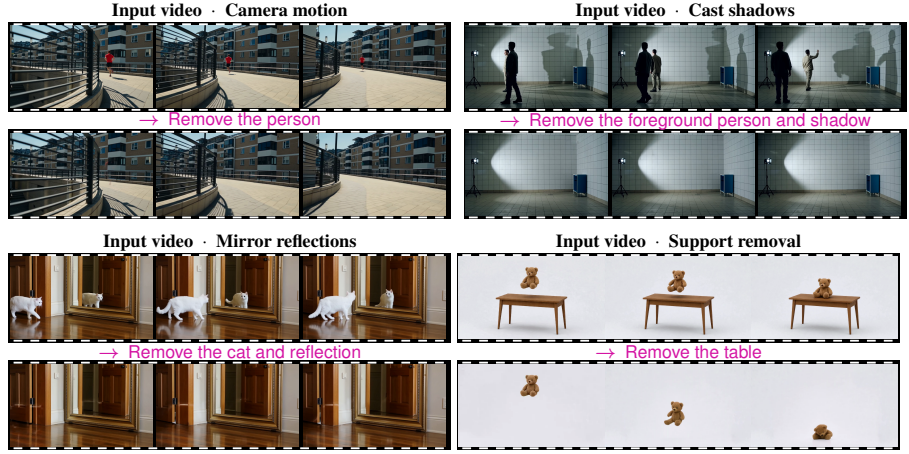}
\end{minipage}
\captionof{figure}{TripleFlow performs high-quality zero-shot video object removal without additional training, producing clean and spatiotemporally coherent results in challenging scenarios. It removes not only the target object, but also associated effects such as cast shadows, mirror reflections, and gravity-induced motion.}
\label{fig:teaser}
\end{center}
\vspace{-0.10in}
\begin{abstract}
Video object removal presents a uniquely difficult editing challenge. Because a removal prompt specifies only what to erase rather than what to generate, the model must infer and reconstruct a highly specific occluded background entirely from the surrounding context. Existing training-free methods struggle with this because their editing mechanisms act primarily as localized erasers. They fail to actively synthesize the missing background details and often leave behind ghosting artifacts. To solve this, we propose \method, a training-free framework that tightly couples erasure and generation. It coordinates a source flow, a residual flow, and a synthesis flow throughout the entire process. By reusing a single target prediction, the residual flow isolates and suppresses the object, while the synthesis flow independently reconstructs the occluded background. Crucially, \method\ injects this newly synthesized background back into the editing trajectory at every step. This continuous feedback loop ensures that the generated structures actively guide the removal process, achieving seamless completion that is spatiotemporally consistent with the unedited scene. Extensive evaluations across five challenging benchmarks demonstrate that \method\ establishes a new state-of-the-art, significantly outperforming existing baselines in both reconstruction fidelity and temporal consistency. 
\end{abstract}
\raggedbottom
\section{Introduction}

\begin{figure}[t]
\centering
\includegraphics[width=\linewidth]{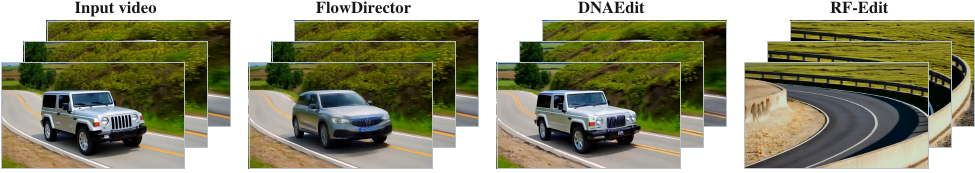}
\caption{\textbf{Object removal with general-purpose video editors.}
Given an empty-road prompt, FlowDirector~\citep{flowdirector} and DNAEdit~\citep{dnaedit} retain vehicles, while RF-Edit~\citep{rfedit} removes the car but substantially changes the road and surroundings.}
\label{fig:editing_vs_removal}
\end{figure}

Video diffusion models and continuous flow formulations have shown strong capabilities in synthesizing realistic motion dynamics and complex visual scenes \citep{videodiffusion,wan2025,cogvideox}. When adapted for video editing, these models naturally excel at semantic replacement, such as turning a running dog into a cat~\citep{tokenflow,flatten}. In these tasks, an explicit positive prompt guides the generation of a new object that naturally covers the missing background. This replacement process also offers significant generative freedom because the model only needs to synthesize a plausible instance of the target object. For example, if we want to replace a dog with a cat, an explicit prompt tells the model exactly what to create, and almost any realistic cat can serve as the replacement. Video object removal demands far more than changing semantic identity because it completely lacks this generative guidance. When general-purpose video editors attempt this task, they typically either fail to erase the foreground object or severely distort the surrounding scene as shown in Figure~\ref{fig:editing_vs_removal}. Since a removal prompt only specifies what to erase and provides no description of the missing surface, the model naturally does not know what to fill in the exposed void. It must instead infer the exact background entirely from the surrounding video context, but this inferred background is never arbitrary. The surrounding scene strictly limits what can fill the space, and earlier or later frames often reveal the same surface under changing visibility~\citep{e2fgvi,propainter}. The model must therefore reconcile these observations across space and time to reconstruct the specific scene hidden behind the object. While a replacement object can simply hide the background, removal completely exposes it, so any unsuccessful background reconstruction instantly reveals errors in geometry, texture, and shading. Successful removal must therefore do more than remove the target object and its associated effects \citep{rose,effecterase}. It must reconstruct the exposed background faithfully while preserving the surrounding scene.

To reconstruct this hidden background, existing inpainters often rely on task-specific training \citep{propainter, diffueraser}, but these supervised approaches demand extensive compute and custom datasets. Training-free editors offer a flexible alternative by leveraging pretrained video generators directly~\citep{anyv2v,contextflow}. However, these methods must carefully balance preserving the unedited scene with generating new background content. Inversion-based approaches \citep{objectwiper} maintain source alignment through numerical inversion, but this process often introduces drift that corrupts the background. Inversion-free formulations provide a strong foundation. They perform source-relative edits by integrating velocity differences between source and target predictions~\citep{flowedit,flowdirector}. This avoids numerical inversion entirely and protects the unedited background from drift. However, applying this purely differential approach to object removal reveals a critical limitation: subtraction is not completion. Integrating velocity differences successfully suppresses the foreground object, but it fails to reconstruct the complex structures hidden behind it. This differential update essentially acts as a localized erasure mechanism. However, it lacks the continuous generative momentum required to create entirely new background content, and ultimately leaves behind faint ghosting artifacts or blurry areas. Previous methods attempt to fix this issue by abruptly transitioning to target-only generation at the final step \citep{flowedit}. This late-stage switch disconnects the synthesis process from the editing trajectory. To achieve complete removal, the framework must integrate generation and erasure from the very beginning. It must continuously synthesize the missing background without sacrificing the precise source-relative control of a differential update.

To this end, we present \textbf{\method}. This inversion-free framework fundamentally redefines object removal by coordinating a \emph{source flow}, a \emph{residual flow}, and a \emph{synthesis flow} throughout the entire sampling process. The source flow acts as an analytic reference trajectory that anchors the edit strictly to the original video. To achieve complete removal without requiring an extra synthesis model, \method\ smartly reuses a single target-velocity prediction to drive two different updates. The residual flow computes a velocity difference to suppress the foreground object and extract the editing direction. Concurrently, the synthesis flow accumulates the same target velocity directly, preserving the pure generative momentum needed to synthesize the missing background. Crucially, these trajectories do not simply fuse at the very end. Instead, \method\ merges the synthesis state into the residual flow at every single noise step. This means the newly generated background is continuously fed back into the ongoing edit. Because the model always predicts its next step based on this merged result, the background synthesis actively steers the editing trajectory. Rather than acting as a final patch, generation becomes a persistent feedback loop that guides the entire removal process. 
Our main contributions are summarized below:
\begin{itemize}
    \item We identify two fundamental challenges of video object removal: the lack of explicit text descriptions for the missing regions, and the uniqueness of the required background, which must strictly align with visual cues revealed in surrounding frames.
    
    \item We propose \textbf{\method}, an inversion-free and training-free framework that integrates erasure and generation into a persistent feedback loop. By coordinating a source, residual, and synthesis flow, it reuses a single target prediction to actively steer the editing trajectory, synthesizing complex missing structures while strictly anchoring the unedited scene.
    
    \item Extensive evaluations across five challenging video benchmarks~\citep{davis,objectwiper,rose,prove} demonstrate that \method\ achieves state-of-the-art performance on all of them, outperforming all existing training-free baselines. Beyond accurate background reconstruction, our framework naturally eliminates associated physical effects such as shadows and reflections. Crucially, our ablation studies validate the distinct and complementary roles of each flow demonstrating that joint flow coordination is essential for eliminating ghosting artifacts and robustly reconstructing the occluded background.

\end{itemize}

\flushbottom
\section{Related Work}

\paragraph{Video inpainting and object removal.}
Video inpainting traditionally propagates spatial and temporal features to reconstruct missing regions \citep{sttn,fuseformer,e2fgvi,propainter}. Recent supervised methods incorporate diffusion priors to hallucinate fine-grained details within the occluded areas \citep{diffueraser}. A complete pipeline must also eliminate object-associated visual effects. Methods like ROSE \citep{rose} and EffectErase \citep{effecterase} explicitly target shadows and reflections using specialized paired training data. These supervised approaches demonstrate the fundamental requirement to faithfully reconstruct the background while completely removing the visual influence of the target object.

\paragraph{Training-free video object removal.}
Recent approaches leverage pretrained video generators to bypass task-specific fine-tuning for zero-shot object removal. OmnimatteZero \citep{omnimattezero} suppresses object-associated effects using point tracking and spatial attention guidance. Object-WIPER \citep{objectwiper} combines effect localization with explicit source inversion and cross-attention manipulation to suppress target identity while preserving the surrounding scene. These paradigms demonstrate the feasibility of controlling generative models by modulating attention and feature retention during the sampling process \citep{prompttoprompt,pnp}.

\paragraph{Flow-based generative editing.}
Continuous flow formulations \citep{flowmatching,rectifiedflow} and diffusion models \citep{ddpm,latentdiffusion} provide a robust foundation for video editing. Inversion-free residual methods enable direct zero-shot editing by operating on velocity predictions. FlowEdit \citep{flowedit} integrates the velocity difference between source and target predictions under shared noise to bypass numerical inversion. FlowDirector \citep{flowdirector} extends this with direction-aware correction and differential averaging guidance. To precisely restrict edit regions, various methods utilize regional attention control \citep{objectwiper}, soft localization \citep{flowdirector}, or temporal mask union \citep{svor}. These flow-based methods excel at semantic replacement by modifying the generative trajectory.
\section{Method}
\label{sec:method}

\begin{figure}[t]
\centering
\includegraphics[width=\linewidth]{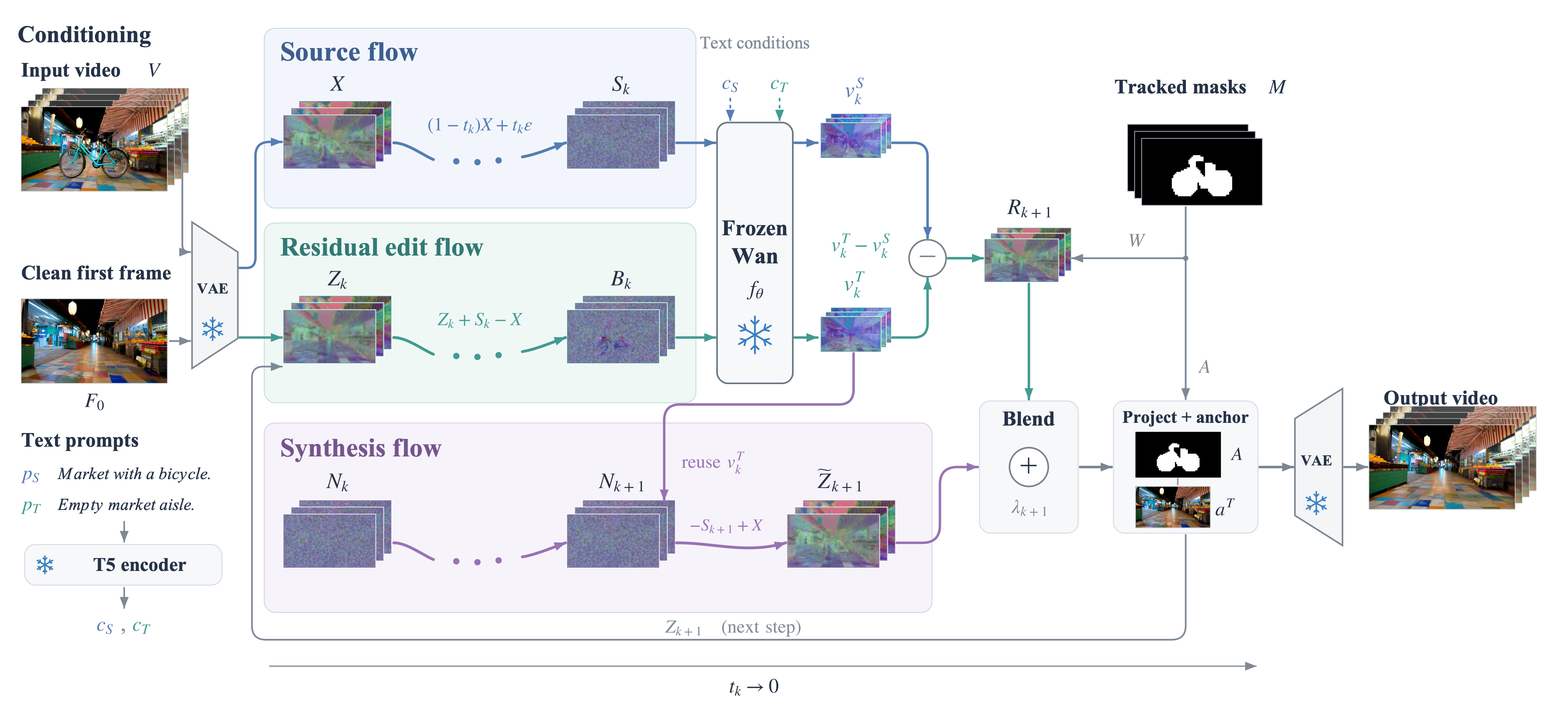}
\caption{\textbf{TripleFlow pipeline.} Our framework coordinates three coupled flows within a frozen video diffusion backbone at each noise step. The source flow preserves the observed scene. By reusing a shared target prediction, the residual flow suppresses the foreground object using a velocity difference while the synthesis flow independently reconstructs the occluded background. The framework injects this synthesized background back into the residual editing trajectory, followed by mask projection before the next step.}
\label{fig:pipeline}
\end{figure}

\subsection{Problem Setup}
\label{sec:preliminaries}
\label{sec:overview}

Given a source video \(V\), a frame-wise removal mask \(M\), and a clean reference first frame \(F_0\), our goal is to remove the masked object and any associated effects covered by \(M\) while preserving the visible scene and its motion. Newly revealed regions should contain plausible, temporally coherent background. The reference \(F_0\), taken from an unoccluded frame or generated by an off-the-shelf image model, provides the desired background appearance in the first frame.

We perform the edit entirely at inference time and keep both the video generator and its causal VAE frozen. Let \(\mathcal{E}\) and \(\mathcal{D}\) denote the VAE encoder and decoder. We use \(X\) for the encoded source video, and \(a^S\) and \(a^T\) for the encoded source and clean first frames. The frozen flow model \(f_\theta\) predicts a velocity from a latent, its noise level \(t\), a text prompt \(p\), and a first-frame condition \(a\). We integrate over \(t_1=1>\cdots>t_K=0\), with \(\Delta t_k=t_{k+1}-t_k<0\). Rather than numerically inverting \(X\) to obtain a source-aligned noise state, we sample one noise tensor \(\epsilon\) and construct the source trajectory analytically. The mask \(M\) is converted into a hard latent support \(A\) and a soft edit weight \(W\), described in Sec.~\ref{sec:localization}.

Source-relative editing preserves the observed scene but can struggle to recover background that was never visible; direct generation can fill the missing region but may alter the scene's appearance or motion. Editing trajectories that drift from the latent interpolation used to train the flow model can also produce artifacts, motivating a source-aligned reference throughout sampling. TripleFlow couples this reference with a residual edit and a synthesis state for the occluded background, combining their estimates before each subsequent target prediction.

\subsection{TripleFlow Dynamics}
\label{sec:tripleflow}

TripleFlow links a source flow, a residual removal flow,
and a synthesis flow. The source flow places the input
video at each noise level using fixed noise, providing
a reference for the original scene. At each step, the
model predicts a source velocity $v_k^S$ from this
reference and a target velocity $v_k^T$ from the current
edit state under the removal condition. Their difference
updates the residual flow, while $v_k^T$ also advances
the synthesis flow to develop the exposed background.
The two resulting edit estimates are combined after
each step. Their combination guides the next target
prediction and is ultimately decoded into the output
video. Only $v_k^S$ and $v_k^T$ are predicted at each step.

\paragraph{Source Flow as Model Reference.}
Given a fixed noise tensor $\epsilon$, we construct a
source-aligned state at each noise level:
\begin{equation}
S_k=(1-t_k)X+t_k\epsilon.
\label{eq:source_flow}
\end{equation}
The pretrained model learns velocities on interpolations between video latents and noise; drifting from these states can introduce artifacts. We therefore use $S_k$ as a source reference throughout sampling. A fixed $\epsilon$ defines a consistent path for comparing source and target velocities at the same noise level without numerical inversion, grounding the edit in the input structure and motion while the target condition guides background completion.

\paragraph{Residual Edit and Synthesis Flow.}
Let $Z_k$ be the current edit estimate in clean-space
coordinates, with $Z_1=X$. The model expects an input
at noise level $t_k$. We place the edit estimate at
that level by adding the offset of the source state
from the input video:
\begin{equation}
B_k=Z_k+S_k-X.
\label{eq:target_query}
\end{equation}
When $Z_k=X$, the target query $B_k$ equals the source
query $S_k$. As the edit develops, the two queries retain
the same source-path offset while differing in their
current content. This lets us compare their predicted
velocities at the same noise level.

We evaluate the frozen model under the source and target
conditions:
\begin{equation}
v_k^S=f_\theta(S_k,t_k;p_S,a^S),\qquad
v_k^T=f_\theta(B_k,t_k;p_T,a^T).
\label{eq:velocities}
\end{equation}
Text can identify the object to remove but cannot determine the appearance of the background it occludes. The source prompt $p_S$ describes the observed video with the object, while $p_T$ describes the scene after removal. The source first-frame condition $a^S$ provides the observed appearance, and the clean condition $a^T$ guides the target background appearance. The source velocity serves as a reference, while the target velocity drives both the residual edit and the synthesis state.

We apply this difference to the clean-space estimate only
where editing is needed. With the soft edit weight $W$,
the residual estimate is
\begin{equation}
R_{k+1}
=Z_k+\Delta t_k\,W\odot(v_k^T-v_k^S).
\label{eq:residual_flow}
\end{equation}
The source velocity keeps the update tied to the input
video, while $W$ controls its strength near the removal
boundary. This source-relative formulation supports the
preservation of visible structure and motion.

To complement source-relative editing, we introduce
a synthesis state $N_k$ that accumulates target-conditioned
velocities directly. It starts from $N_1=\epsilon$ and
evolves at the current noise level:
\begin{equation}
N_{k+1}=N_k+\Delta t_k\,v_k^T.
\label{eq:synthesis_flow}
\end{equation}
This state retains the effect of earlier target-conditioned
updates as background content forms. It reuses the target
velocity evaluated at $B_k$, so the two updates require
the same pair of velocity predictions, $v_k^S$ and $v_k^T$.

We localize the synthesis state using the hard support
$A$, restoring the source state outside the edit region
after each update:
\begin{equation}
N_{k+1}\leftarrow
A\odot N_{k+1}+(1-A)\odot S_{k+1}.
\label{eq:synthesis_projection}
\end{equation}
Before combining the two estimates, we express $N_{k+1}$
in clean-space coordinates:
\begin{equation}
\widetilde Z_{k+1}=N_{k+1}-S_{k+1}+X.
\label{eq:synthesis_clean}
\end{equation}
This conversion maps the source reference $S_{k+1}$
back to $X$. The differences accumulated by the synthesis
state are thereby carried into $\widetilde Z_{k+1}$
within the edit support.

The residual and synthesis estimates share the target
velocity but evolve relative to different source references.
For an interior edited token in a non-anchor frame,
where $A=W=1$, their difference satisfies
\begin{equation}
\widetilde Z_{k+1}-R_{k+1}
=
\widetilde Z_k-Z_k
+\Delta t_k\bigl[v_k^S-(\epsilon-X)\bigr].
\label{eq:state_difference}
\end{equation}
Here, $\epsilon-X$ is the velocity of the analytic source
path. The residual estimate measures change relative
to the learned source velocity, while the synthesis
estimate, expressed in clean-space coordinates, measures
change relative to this analytic velocity. Their
difference therefore retains the accumulated state
discrepancy and incorporates the current difference
between the two source references.

\paragraph{Stepwise Coupled Blending.}
We combine the residual and synthesis estimates to obtain
the next edit state:
\begin{equation}
\begin{aligned}
Z_{k+1}
&=(1-\lambda_{k+1})R_{k+1}
  +\lambda_{k+1}\widetilde Z_{k+1},\\
\lambda_{k+1}
&=\lambda_{\max}(1-t_{k+1})^\gamma.
\end{aligned}
\label{eq:flow_coupling}
\end{equation}

Here, $\lambda_{\max}$ sets the synthesis weight near the clean endpoint, and $\gamma$ controls how quickly it increases. Early steps favor the residual estimate to preserve source layout and motion; later steps give more weight to background synthesis. At each step, $Z_{k+1}$ forms the next target query $B_{k+1}$, so both estimates influence subsequent velocity predictions.
\paragraph{Spatiotemporal conditioning.}
\label{sec:localization}
Instead of using internal attention maps, we use standard segmentation models (e.g., SAM~2~\cite{sam2}, SAM~3~\cite{sam3}, Grounded SAM~2~\cite{groundedsam2}) to get accurate frame-wise object masks $M$ without extra training.

Following the VAE's temporal layout, the first mask is kept separate. For later latent frames, we combine the masks using a pixelwise union. We resize them to the latent spatial resolution using nearest-neighbor interpolation to form the aligned mask $\overline M$.

This aligned mask defines the hard support and soft weight:
\begin{equation}
A=\operatorname{Dilate}(\overline M),
\qquad
W=A\odot\exp(-\rho d_{\overline M}).
\label{eq:edit_support}
\end{equation}
Here, spatial dilation expands the boundary, $d_{\overline M}$ is the distance to the masked region, and $\rho$ controls how fast the boundary fades. $W$ controls the residual update strength, while $A$ limits where the video can change from the source.

To guide the new background, we use a pretrained image editor $\mathcal G$. A vision-language model finds video frames $\mathcal R$ that show the hidden background. The editor $\mathcal G$ uses the source first frame $V_0$ and these found frames $\{V_j\}_{j\in\mathcal R}$ to make a clean first-frame reference $F_0$:
\begin{equation}
F_0=\mathcal G\!\left(V_0,\{V_j\}_{j\in\mathcal R}\right),
\qquad
a^T=\mathcal E(F_0).
\label{eq:reference_construction}
\end{equation}
The encoded reference $a^T$ then guides every target velocity evaluation.

These spatial and appearance conditions also constrain the state after each step:
\begin{align}
Z_{k+1}
&\leftarrow
A\odot Z_{k+1}+(1-A)\odot X,
\label{eq:source_projection}\\
Z_{k+1}^{(0)}
&\leftarrow
(1-F)\odot X^{(0)}+F\odot a^T,
\qquad F=W^{(0)}.
\label{eq:first_frame_anchor}
\end{align}
Eq.~\eqref{eq:source_projection} keeps the original video outside the mask. Eq.~\eqref{eq:first_frame_anchor} links the first-frame edit region to the clean reference, using $F$ to smooth the transition.
\paragraph{Localized Target Attention Control.}
Spatial masks limit the changes of latent values. However, the target velocity also depends on features from self-attention. During early sampling steps, masked queries can still retrieve object features from masked keys and pass them into the generated content. Therefore, we introduce attention control to complement the spatial masks.

Let $h_i\in\{0,1\}$ indicate whether token $i$ overlaps $\overline M$. We obtain this by max-pooling the aligned mask over the patch grid. Following regional attention scaling~\cite{objectwiper}, we scale the masked keys and apply a negative bias to the masked query--key pairs. For transformer block $b$ at sampling step $k$, the attention logits become:
\begin{equation}
\begin{aligned}
\ell_{ij}^{k,b}
&=
\left[1-(1-\alpha_k)h_j\right]
\frac{\mathbf q_i^\top\mathbf k_j}{\sqrt d}
-\beta_{k,b}h_i h_j,\\
P_{ij}^{k,b}
&=
\operatorname{softmax}_{j}
\left(\ell_{ij}^{k,b}\right),
\end{aligned}
\label{eq:localized_attention}
\end{equation}
where $\mathbf q_i$ and $\mathbf k_j$ are the normalized, position-encoded query and key vectors. During the first quarter of sampling steps, we set $\alpha_k=0.5$ in all blocks and $\beta_{k,b}=5$ in the first ten blocks. Otherwise, $\alpha_k=1$ and $\beta_{k,b}=0$.

Key scaling limits the contribution of masked keys to the attention scores. Meanwhile, the negative bias prevents masked queries from retrieving masked keys. These operations define the controlled target model $f_\theta^{\mathrm{mask}}$, which is used for $v_k^T$ in Eq.~\eqref{eq:velocities}. By sharing this target prediction, we apply the attention control to both the residual and synthesis updates. This effectively complements the spatial support and the clean reference.

\section{Experiments}
\label{sec:experiments}

\noindent\textbf{Datasets and Baselines.}
To evaluate our method, we conduct experiments on four video object removal benchmarks, including DAVIS~\citep{davis}, WIPER-Bench~\citep{objectwiper}, ROSE~\citep{rose}, and PROVE~\citep{prove}. These benchmarks encompass diverse scenarios with various subjects and dynamic camera motions, alongside complex object-associated effects such as shadows and reflections. Specifically, ROSE and PROVE-M offer paired clean ground truths to assess background reconstruction, whereas PROVE-H comprises challenging in-the-wild videos without paired references. We compare our approach with three representative training-free video editing methods: OmnimatteZero~\citep{omnimattezero}, Object-WIPER~\citep{objectwiper}, and ContextFlow~\citep{contextflow}. We additionally evaluate OmniEraser~\citep{omnieraser}, a trained image object removal model applied independently to video frames. For quantitative evaluation, we report CORE~\citep{core} to evaluate object and associated-effect removal, along with PSNR~\citep{psnr}, SSIM~\citep{ssim}, and LPIPS~\citep{lpips} on paired datasets to measure reconstruction quality.

\label{sec:evaluation}

\begin{figure}[tbp]
\centering
\includegraphics[width=0.9\linewidth]{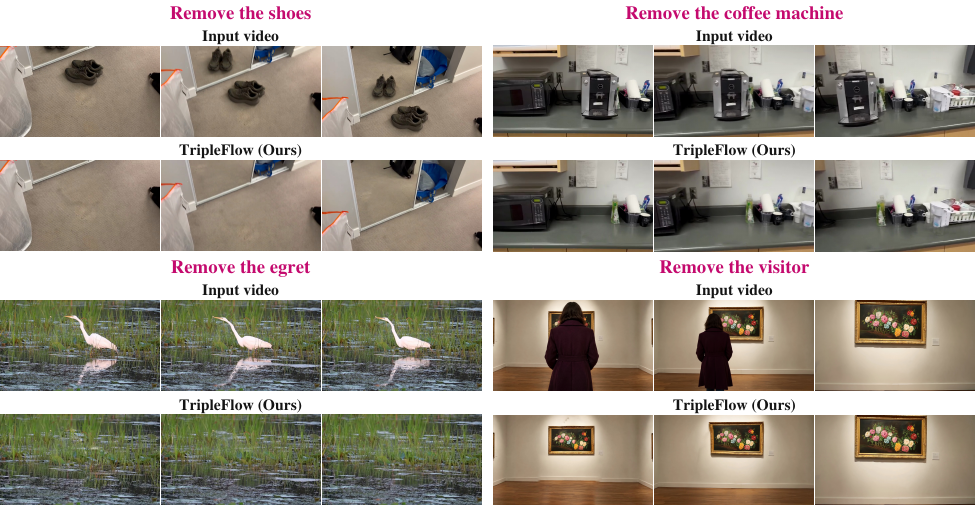}
\caption{\textbf{Video object removal under camera motion.}
TripleFlow removes foreground subjects and reconstructs temporally coherent backgrounds while preserving the original camera motion and scene structure. }
\label{fig:qualitative_results}
\end{figure}

\begin{figure}[tbp]
\centering
\includegraphics[width=0.9\linewidth]{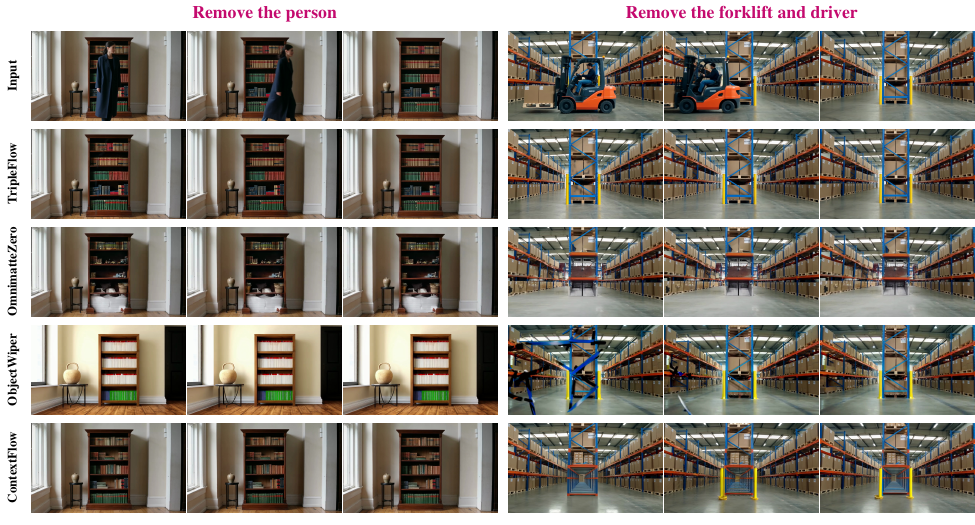}
\caption{\textbf{Qualitative comparison with existing video object removal methods.}
TripleFlow achieves clean removal while preserving fine background details and scene geometry, as illustrated by the bookshelf contents and warehouse shelving. In contrast, OmnimatteZero, ObjectWiper, and ContextFlow exhibit residual objects, altered background appearance, or structural distortions in these examples.}
\label{fig:qualitative_comparison}
\end{figure}

\noindent\textbf{Qualitative Evaluation.}
\method\ faithfully reconstructs backgrounds that are initially occluded but revealed later in the video (Fig.~\ref{fig:qualitative_results}). It recovers the painting behind the gallery visitor and the countertop behind the coffee machine, maintaining visual consistency with their later appearances. In the gallery sequence, the painting retains its placement as the camera moves rather than being replaced by an arbitrary wall. By contrast, the baselines in Fig.~\ref{fig:qualitative_comparison} remove the foreground person and forklift but fail to preserve surrounding structural details within the masked regions. In the bookcase sequence, OmnimatteZero, ObjectWiper, and ContextFlow~\citep{contextflow} alter shelf contents or introduce spurious horizontal structures; ObjectWiper also distorts the surrounding floor and wall. In the warehouse, the baselines distort or fabricate the central blue-and-yellow rack. \method\ instead preserves the book arrangement and shelving geometry. Additional qualitative results and baseline comparisons are provided in Appendices~\ref{app:qualitative_results} and~\ref{app:qualitative_comparisons}, respectively.

Beyond background reconstruction, \method\ removes associated physical effects without effect-specific training, including the shoes' mirror reflection and the egret's water reflection (Fig.~\ref{fig:qualitative_results}). This capability follows directly from maintaining separate editing and synthesis flows. Once the foreground is suppressed within the masked region, the synthesis state decouples from the object's visual characteristics and relies on the unedited residual flow from the surroundings to propagate clean background content. The completed regions consequently remain structurally consistent with the bare floor or empty water surface.

\noindent\textbf{Quantitative Evaluation.}
\method\ achieves the best mean CORE across all five benchmarks (Table~\ref{tab:core_benchmarks}) and the best PSNR, SSIM, and LPIPS on both paired benchmarks (Table~\ref{tab:reconstruction_metrics}). Its PSNR exceeds the strongest competing results by nearly 4~dB on PROVE-M and 2~dB on ROSE. On PROVE-M, OmniEraser obtains a similar CORE score (3.125 versus 3.154) but nearly twice the LPIPS (0.3407 versus 0.1724), showing that similar removal scores can conceal substantial differences in background fidelity.

In a five-method human evaluation on ten showcase videos, \method\ receives a 56.7\% first-place preference rate, compared with 23.3\% for the runner-up, OmnimatteZero (Fig.~\ref{fig:radar_metrics}, middle. See Appendix~\ref{app:human_evaluation} for the study setup).. This preference is consistent with the reduced visual artifacts observed in the baselines. On pooled PROVE-M and ROSE results, \method\ has both the lowest masked-region LPIPS and the highest outside-mask PSNR (Fig.~\ref{fig:radar_metrics}, left), indicating stronger completion without sacrificing the surrounding scene. It also achieves the best mean tLP and tOF~\citep{tecogan} scores. Finally, \method\ attains a higher mean CORE while running significantly faster than flow-based editing baselines such as ContextFlow (Fig.~\ref{fig:radar_metrics}, right). Implementation and evaluation details are provided in the appendix.

\begin{table}[!htbp]
\centering
\caption{Video object removal results. CORE is computed as the mean of ObjectScore and AftereffectScore (higher is better).}
\label{tab:core_benchmarks}
\small
\setlength{\tabcolsep}{6pt}
\begin{tabular}{lccccc}
\toprule
Method & DAVIS $\uparrow$ & WIPER $\uparrow$ & PROVE-M $\uparrow$ & PROVE-H $\uparrow$ & ROSE $\uparrow$ \\
\midrule
OmnimatteZero & 2.917 & 3.188 & 2.576 & 3.167 & 3.260 \\
ObjectWiper & 2.056 & 2.375 & 2.232 & 2.389 & 2.625 \\
ContextFlow & 2.667 & 3.500 & 2.438 & 3.111 & 2.357 \\
OmniEraser & 3.000 & 3.188 & 3.125 & 2.889 & 2.929 \\
\midrule
\method\ (Ours) & \textbf{3.394} & \textbf{3.576} & \textbf{3.154} & \textbf{3.275} & \textbf{3.325} \\
\bottomrule
\end{tabular}

\end{table}

\begin{table}[!htbp]
\centering
\caption{Reconstruction quality against ground truth labels on PROVE-M and ROSE.}
\label{tab:reconstruction_metrics}
\small
\setlength{\tabcolsep}{5pt}
\begin{tabular}{lcccccc}
\toprule
\multirow{2}{*}{Method} & \multicolumn{3}{c}{PROVE-M} & \multicolumn{3}{c}{ROSE} \\
\cmidrule(lr){2-4}\cmidrule(lr){5-7}
 & PSNR $\uparrow$ & SSIM $\uparrow$ & LPIPS $\downarrow$ & PSNR $\uparrow$ & SSIM $\uparrow$ & LPIPS $\downarrow$ \\
\midrule
\method\ & \textbf{24.636} & \textbf{0.8644} & \textbf{0.1724} & \textbf{27.387} & \textbf{0.9112} & \textbf{0.1053} \\
OmnimatteZero & 20.894 & 0.8106 & 0.2819 & 25.450 & 0.8687 & 0.1877 \\
ObjectWiper & 16.608 & 0.6616 & 0.4129 & 18.897 & 0.6893 & 0.3655 \\
ContextFlow & 17.951 & 0.7912 & 0.3144 & 22.015 & 0.8912 & 0.1599 \\
OmniEraser & 19.044 & 0.7881 & 0.3407 & 19.649 & 0.7848 & 0.2741 \\
\bottomrule
\end{tabular}
\end{table}

\begin{figure}[!htbp]
\centering
\includegraphics[width=0.7\linewidth]{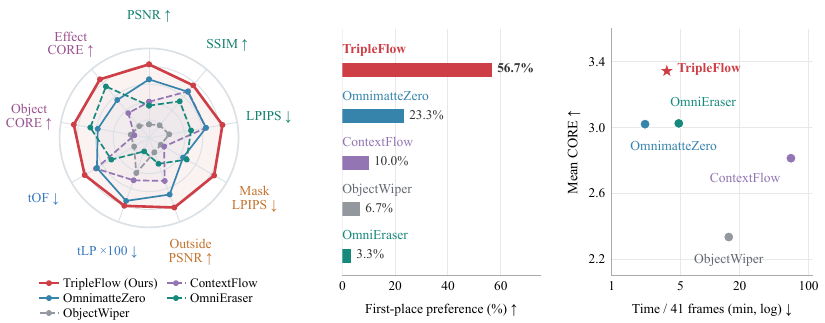}
\caption{
\textbf{Left:} Radar comparison across nine combined metrics from PROVE-M and ROSE. Further outward indicates better performance.
\textbf{Middle:} First-place preference rates in the five-method human evaluation on ten showcase videos.
\textbf{Right:} Performance-speed trade-off.}
\label{fig:radar_metrics}
\label{fig:runtime_comparison}
\label{fig:human_evaluation}
\end{figure}

\begin{figure}[!htbp]
    \centering
    \includegraphics[width= 0.9\linewidth]{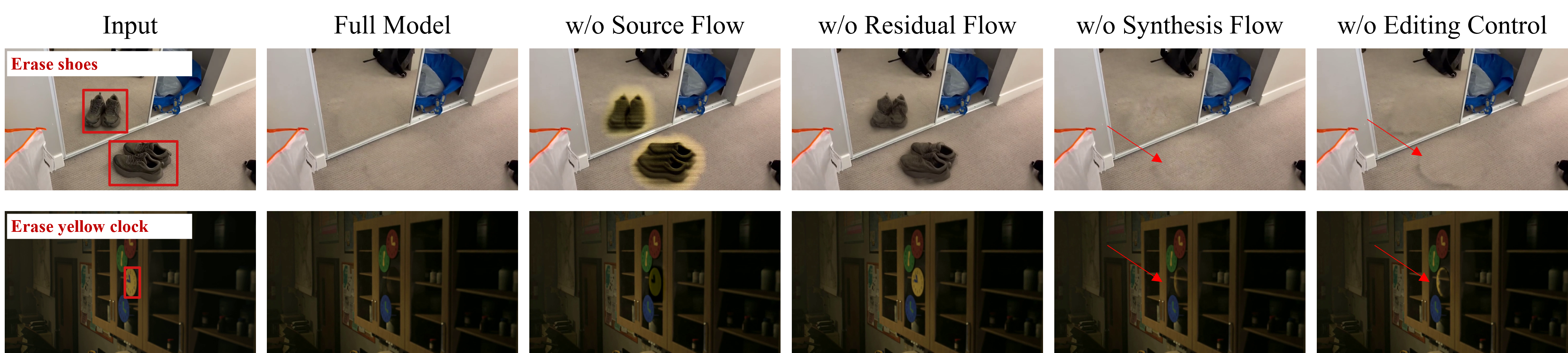}
    \caption{\textbf{Qualitative ablation results.} The input frames mark the removal targets: shoes and their mirror reflection (top), and a yellow clock (bottom). Full \method removes the targets while preserving the surrounding scene. Ablating individual flows or editing control can leave object remnants or introduce visible artifacts.}
    \label{fig:ablation}
\end{figure}

\noindent\textbf{Ablation Study.}
We ablate the flows and control mechanisms of \method\ on three representative cases (Fig.~\ref{fig:ablation}). Replacing \(v_{\mathrm{tar}}-v_{\mathrm{src}}\) with the target velocity alone (w/o Source Flow) introduces object-shaped artifacts and appearance distortion, indicating drift from the observed scene. Suppressing the residual update (w/o Residual Flow) leaves the objects largely exist, confirming its role in removal. Without native-state synthesis (w/o Synthesis Flow), removal remains possible but background completion and fine details become less reliable. Removing Editing Control introduces localized leakage and boundary artifacts and working together with Synthesis Flow for detailed boundary remove quality. Quantitative and more qualitative results in Appendix~\ref{app:ablation} Table~\ref{tab:n7_ablation_20pct} and Fig.~\ref{fig:app_ablation_qualitative}, both indicate that full model consistently outperforms all four ablations.

\FloatBarrier

\section{Conclusion}
\label{sec:conclusion}
We propose \method, a training-free video object removal framework that coordinates a source, a residual, and a synthesis flow within a single diffusion backbone. By injecting the synthesized background into the residual trajectory, our approach seamlessly couples target erasure with active scene reconstruction. Evaluations across five benchmarks demonstrate \method\ significantly outperforms baselines in reconstruction fidelity, temporal consistency, and the natural elimination of associated physical effects.

\subsection*{Reproducibility Statement}

We describe the formulation and inference procedure of \method{}
in Section~\ref{sec:method}.
Appendix~\ref{app:implementation} provides the pretrained backbone,
sampling hyperparameters, mask preparation, clean first-frame
generation, and baseline configurations.
The benchmarks and evaluation metrics are described in
Section~\ref{sec:experiments}, while
Appendix~\ref{app:ablation} specifies the evaluation subsets and
aggregation protocol for the additional ablation studies.
The supplementary materials include our core implementation,
inference configuration, a runnable example, and video results
to facilitate reproduction and inspection of temporal behavior.
\bibliography{references}
\clearpage
\appendix
\section{Implementation Details}
\label{sec:appendix}
\label{app:implementation}

\subsection{Backbone and Sampling}
\label{app:sampling}

We implement \method{} with the pretrained Wan2.2-TI2V-5B backbone~\citep{wan22}.
The video transformer, text encoder, and causal VAE remain frozen throughout
inference. We use 40 sampling steps with a timestep shift of 12 and initialize
the random generator with seed 42. The source and target classifier-free
guidance scales are 3.5 and 5.0, respectively. The source path uses a single
Gaussian noise realization shared across all sampling steps; numerical source
inversion and multi-noise averaging are not used.

The target query uses $B_k=Z_k+S_k-X$, with query coupling set to zero.
Updated states are combined using $\lambda_{k+1}$, where
$\lambda_k=0.3(1-t_k)^{1.5}$; synthesis influences the next query through
$Z_{k+1}$. Each step computes two guided velocities, requiring four
transformer forward passes for the conditional and negative-prompt
evaluations. The synthesis update reuses the target velocity without
additional backbone evaluations.

We process each video jointly in a single temporal window. For a clip of
$L$ frames, we repeat the final source frame and its mask until the length is
$L'=1+4\lceil(L-1)/4\rceil$, as required by the causal VAE, and discard only
these padding frames after decoding. All original frames and their timing are
retained. Spatial preprocessing uses the backbone's $1280\times704$ maximum-area
setting, with input-dependent dimensions aligned to its latent and patch grids.
The source video, masks, and first-frame conditions undergo corresponding
spatial transforms; exported videos are restored to the source dimensions.

\subsection{Appearance and Spatial Conditioning}
\label{app:conditioning}

\paragraph{Scene descriptions and clean reference.}
The source caption describes the observed scene; the target caption describes
the same scene after removal, retaining its background layout, camera motion,
and remaining objects. The source and target branches use the original and
clean first frames, respectively.

We generate the clean first frame with Image2.5, conditioned on the original
first frame, its target-region mask, and one to three later source frames
selected for their visibility of the occluded background. The editing
instruction requests removal of the target and its associated effects while
preserving the first frame's viewpoint, lighting, and unrelated content.
No clean ground truth is supplied. Only the final edited image
conditions Wan; the reference images are not separately injected during
sampling. We retain the original first frame when the target and its effects
are already absent. The evaluated collection also includes recorded Image2
fallbacks where an Image2.5 reference was unavailable.

\paragraph{Mask preparation and latent support.}
We obtain frame-wise masks with SAM~3~\citep{sam3} and take the union of the
target-object, shadow, and reflection regions as the editing mask. Target
identities follow the benchmark annotations. Masks are visually checked and
manually corrected against the source video for substantial omissions or
over-segmentation. Associated effects to be removed are included in the
editing support.

The first RGB mask maps to the first latent slice, while each subsequent
group of four masks is combined by a temporal union. We resize masks to the
latent grid with nearest-neighbor interpolation and binarize them at 0.2.
The hard support $A$ uses a $3\times3$ spatial dilation of the aligned binary
mask. The soft weight in Eq.~\eqref{eq:edit_support} uses $\rho=0.5$ and
distance measured in latent-grid pixels. No additional input-mask dilation
is applied by the N7 sampling adapter. After each coupled update, we project
the edit state back to $X$ outside $A$ and anchor its first latent slice with
the clean reference using $W^{(0)}$, as in
Eqs.~\eqref{eq:source_projection}--\eqref{eq:first_frame_anchor}.

\paragraph{First-frame queries and attention control.}
Before each velocity evaluation, the first latent slice is replaced by the
corresponding source or target first-frame condition, and its token timesteps
are set to zero. Regional self-attention control is active only in the target
branch during the first 10 of the 40 sampling steps. Masked keys are scaled by
0.5 in every transformer block; an additional logit bias of $-5$ is applied
to masked-query/masked-key pairs in the first 10 blocks. Both operations are
applied to the conditional and negative-prompt target evaluations. Other
steps and the source branch use the unmodified attention computation.

\subsection{Baseline Implementations}
\label{app:baseline_implementations}

\paragraph{OmnimatteZero.}
Our local OmnimatteZero~\citep{omnimattezero} runs use its released LTX-Video-0.9.7
implementation with 30 inference steps and seed 42. The source video and mask
sequence are passed as video conditions, with the prompt \texttt{Empty} and
the negative prompt \texttt{worst quality, inconsistent motion, blurry, jittery,
distorted}. These runs do not receive our generated clean first frames.
Spatial dimensions are aligned to multiples of 32, and temporal padding is
removed before restoring the output to the source dimensions and frame count.

\paragraph{Object-WIPER.}
Our local Object-WIPER~\citep{objectwiper} runs use the original HunyuanVideo
backbone with 25 rectified-flow steps, flow shift 7, seed 42, and guidance
scales of 1.0 for inversion and 5.0 for denoising. We supply the source and
object-removed scene captions and the prepared union mask. The implementation
retains its associated-effect localization, foreground noise reinitialization,
and background value-feature reuse. Its internal mask dilation is set to 9.
CPU offloading is enabled for the large backbone.

\paragraph{ContextFlow.}
The ContextFlow~\citep{contextflow} quality results use the original Wan2.1-I2V-14B
backbone. We use 50 midpoint rectified-flow inversion intervals and 50 editing
intervals, timestep shift 5, guidance scale 3, and seed 42. Source key/value
features are injected during the first 25 editing intervals. ContextFlow's clean first frame is generated independently with it official MagicQuill
from the original first frame, its reviewed union mask, and the same
scene-specific target caption. MagicQuill uses 20 steps, guidance scale 5,
seed 42, the Euler ancestral sampler, and the Karras schedule. It receives
an empty negative prompt, mask growth of 15, inpainting strength 1.0, color
strength 0.55, and edge strength $0.55/3$. It receives neither our generated
first frame nor later reference images. The mask
guides this image-editing stage; the subsequent ContextFlow video stage
does not take an explicit object mask.

\paragraph{OmniEraser.}
We apply the released OmniEraser Base model~\citep{omnieraser}, consisting of
FLUX.1-dev and its trained removal weights, independently to every video
frame. Quality results use $512\times512$ inputs, 28 inference steps, guidance
scale 3.5, seed 24 reset for each frame, and the prompt
\texttt{There is nothing here.} We use the original benchmark masks without
manual repairs, effect expansion, or dilation. No generated first-frame
condition or temporal state is shared between frames. The model runs in
BF16 with VAE tiling and full GPU residency, and its outputs are resized
back to the source dimensions before video assembly.

\paragraph{Execution and runtime configurations.}
Each \method{} video is processed on one GPU, with independent clips assigned
to different GPUs. Our runs use A100, H200, or RTX 4090 D GPUs, with model
offloading where needed. The runtime plot in Fig.~\ref{fig:runtime_comparison}
combines configuration-specific records: ContextFlow uses the earlier 14B
run on two RTX 4090 D GPUs; OmnimatteZero uses the authors' reported Wan2.1
rate; and OmniEraser uses a separate $1024\times1024$ H200 timing probe
extrapolated to 41 frames. The historical ContextFlow timing run also used a
shared generated first-frame reference, whereas the reported 5B quality
results use MagicQuill. The measured \method{}, Object-WIPER, and ContextFlow
elapsed times include model loading, inference, and export, but exclude
first-frame and mask preparation. The OmniEraser probe excludes loading,
warm-up, and file output. The plot therefore compares the recorded
implementations rather than controlled, same-hardware end-to-end runtimes.

\subsection{Human Evaluation}
\label{app:human_evaluation}

\paragraph{Study materials.}
We conduct a large-scale pilot preference study on ten selected synthetic
showcase videos generated with Wan. Each case contains the source video and
five edited results from \method{}, OmnimatteZero, Object-WIPER, ContextFlow,
and OmniEraser, giving 50 edited videos in total. All displayed clips contain
41 frames at 16 fps and a resolution of $832\times480$, including the first
frame. The ContextFlow videos in this study use the stored 5B configuration
with independently generated MagicQuill first frames; the OmniEraser videos
use frame-wise $512\times512$ inference followed by restoration to the source
resolution. These are selected showcase cases rather than a random sample
from the public benchmarks, so the preference results apply to this set.

\paragraph{Presentation and ranking task.}
The browser-based questionnaire presents a source video, a scene-specific
removal instruction in Chinese/English, and the five results for each case. The
instruction identifies the target and any shadows or reflections to remove,
while specifying background content and camera motion to preserve. Method
names are hidden and replaced with labels A--E. Case order is randomized for
each questionnaire, and the method-to-label assignment is independently
randomized within each case; assignments remain fixed when a session is
resumed. The interface supports synchronized playback, pausing, and replay.
Viewers are instructed to watch the complete clips and rank all five results
from best to worst, with no ties. They jointly consider removal completeness,
background naturalness and preservation of non-target content, and temporal
coherence, including flicker, jitter, and abrupt changes. Each ranking thus
expresses an overall preference across these criteria.

\paragraph{Participation and recording.}
Participation is voluntary and begins with an explicit consent checkbox.
The questionnaire records anonymous rankings and response times without
requesting names, telephone numbers, or email addresses. Responses can be
revised before final submission, which requires completing all ten cases.
The snapshot used for Fig.~\ref{fig:human_evaluation} contains nine completed
anonymous questionnaires, yielding 200 case-level rankings. These are
browser-session records; the identities of respondents were not verified.

\paragraph{Aggregation.}
We use only completed five-method questionnaires, excluding incomplete
responses, developer test submissions, and the earlier four-method protocol.
No response-time threshold is applied. The stored method-to-label assignments
are decoded before aggregation. For each method, the first-place preference
rate is the number of case-level rankings that place it first divided by
90, multiplied by 100. All case-level rankings receive equal weight; because
every completed questionnaire covers all ten cases, each case also receives
the same number of rankings. The reported rates of $56.7\%$ for \method{}
and $23.3\%$ for OmnimatteZero correspond to 51 and 21 first-place rankings,
respectively. The 200 rankings include repeated judgments from the same
questionnaire and are not 200 independent respondents.

\clearpage
\section{Additional Ablation Results}
\label{app:ablation}

\begin{table}[!htbp]
\centering
\small
\setlength{\tabcolsep}{3pt}
\caption{\textbf{N7 ablations on the fixed 20\% evaluation subsets.}
All methods are evaluated on the same 16 PROVE-M and 12 ROSE cases.
PSNR, SSIM, LPIPS, and hole-region PSNR are averaged per video
over frames 1 onward, then equally averaged across cases.
CORE-O and CORE-A denote paired CORE object-removal and aftereffect
scores, respectively. Higher is better except for LPIPS.
The \emph{w/o Editing Control} variant retains mask projection and
first-frame anchoring.}
\label{tab:n7_ablation_20pct}
\begin{tabular}{lcccccc}
\toprule
Method
& PSNR $\uparrow$
& SSIM $\uparrow$
& LPIPS $\downarrow$
& Hole PSNR $\uparrow$
& CORE-O $\uparrow$
& CORE-A $\uparrow$ \\
\midrule
\multicolumn{7}{l}{\textit{PROVE-M (16 cases)}} \\
Full N7
& \textbf{22.57} & \textbf{0.840} & \textbf{0.162}
& \textbf{19.24} & \textbf{2.69} & \textbf{2.50} \\
w/o Source Flow
& 19.33 & 0.778 & 0.213 & 9.60 & 2.00 & 2.06 \\
w/o Residual Flow
& 20.11 & 0.808 & 0.198 & 10.43 & 2.00 & 2.00 \\
w/o Synthesis Flow
& 22.28 & 0.834 & 0.170 & 17.34 & 2.44 & 2.31 \\
w/o Editing Control
& 21.54 & 0.824 & 0.181 & 15.13 & 2.19 & 2.06 \\
\midrule
\multicolumn{7}{l}{\textit{ROSE (12 cases)}} \\
Full N7
& \textbf{24.27} & \textbf{0.859} & \textbf{0.148}
& \textbf{19.79} & \textbf{3.08} & \textbf{2.92} \\
w/o Source Flow
& 19.81 & 0.809 & 0.198 & 10.27 & 2.00 & 2.25 \\
w/o Residual Flow
& 21.47 & 0.837 & 0.184 & 12.56 & 2.00 & 2.17 \\
w/o Synthesis Flow
& 24.06 & 0.857 & 0.152 & 18.90 & 2.75 & 2.83 \\
w/o Editing Control
& 23.58 & 0.854 & 0.157 & 18.06 & 2.33 & 2.50 \\
\bottomrule
\end{tabular}
\end{table}
\FloatBarrier

% Additional qualitative ablations supplied by the authors.
\clearpage
\begin{figure}[p]
\centering
\includegraphics[width=\linewidth,height=0.88\textheight,keepaspectratio]{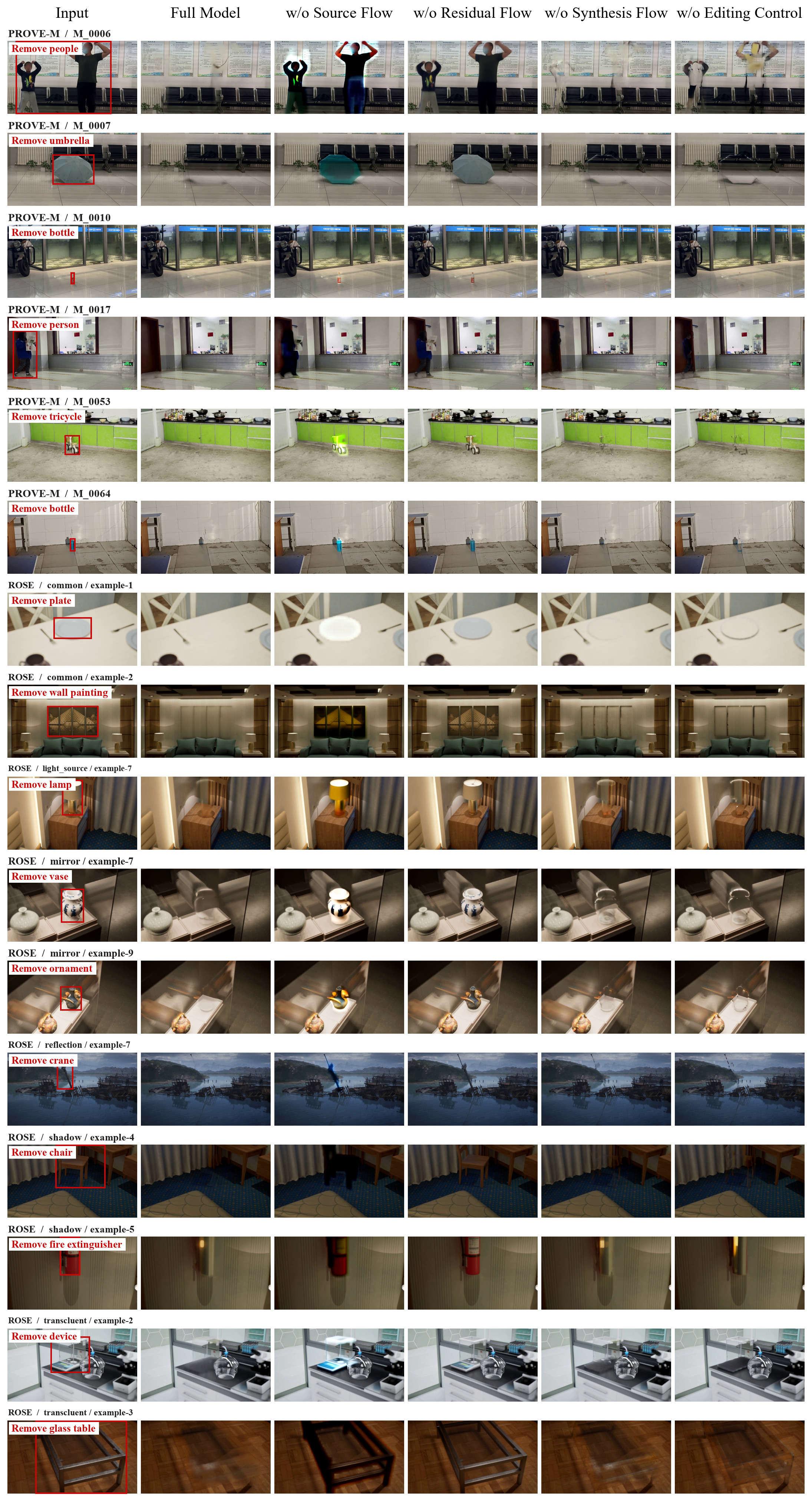}
\caption{\textbf{Additional qualitative ablations on PROVE-M and ROSE.}
Each row compares the input, full model, and variants without the source flow, residual flow, synthesis flow, or editing control. Red boxes indicate the removal targets.}
\label{fig:app_ablation_qualitative}
\end{figure}

% BEGIN SIX-FRAME QUALITATIVE APPENDIX
\clearpage
\section{Additional Qualitative Results}
\label{app:qualitative_results}
We present 10 additional examples from synthetic scenes, captured videos,
and existing benchmarks. Each figure shows the input video above the
\method{} result. Six corresponding frames span the clip from its first
to its last frame, in temporal order from left to right.

\begin{figure}[!htbp]
\centering
% me12_kayak_canal
\includegraphics[width=\linewidth]{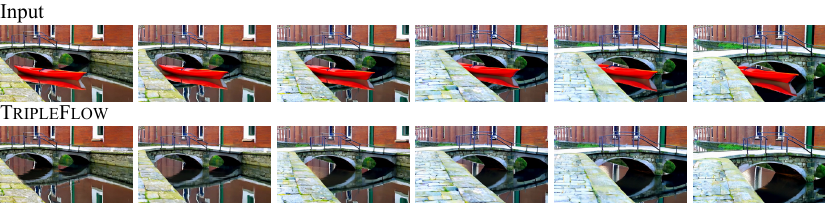}
\caption{Remove the kayak from the canal.}
\label{fig:app-me12-kayak-canal}
\end{figure}

\begin{figure}[!htbp]
\centering
% egret_2_1
\includegraphics[width=\linewidth]{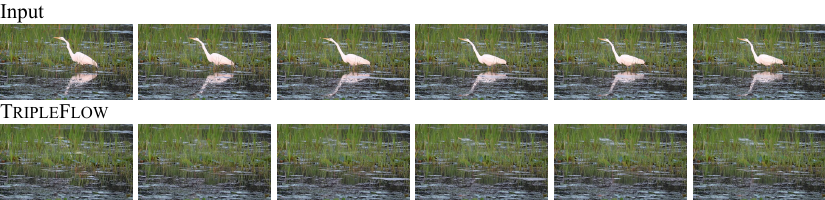}
\caption{Remove the egret from the wetland.}
\label{fig:app-egret-2-1}
\end{figure}

\begin{figure}[!htbp]
\centering
% wild_mirror_hard
\includegraphics[width=\linewidth]{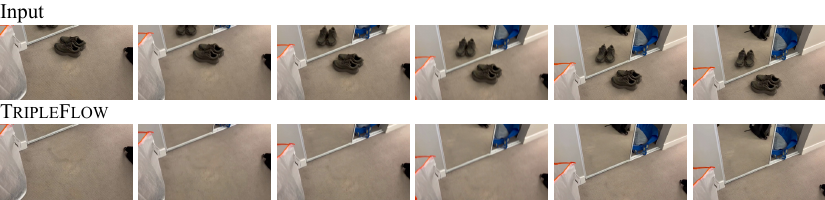}
\caption{Remove the shoes in front of the mirror (hard case).}
\label{fig:app-wild-mirror-hard}
\end{figure}

\begin{figure}[!htbp]
\centering
% wild_kitchen
\includegraphics[width=\linewidth]{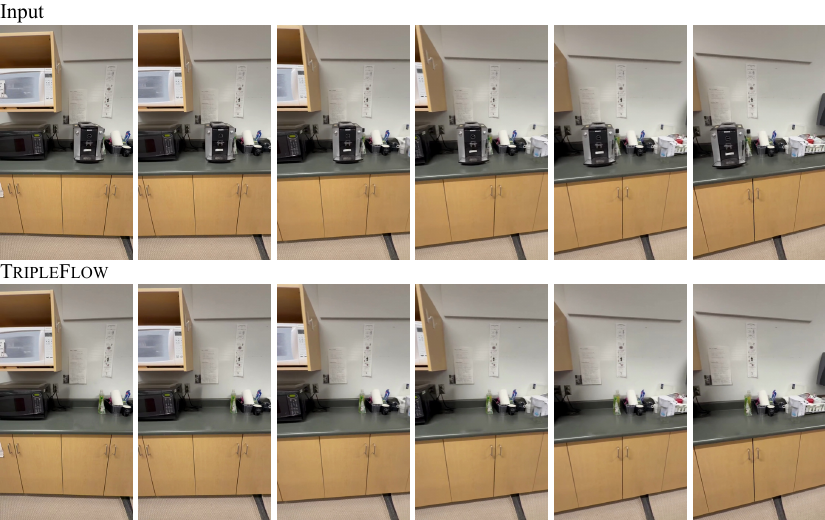}
\caption{Remove the coffee machine from the kitchen counter.}
\label{fig:app-wild-kitchen}
\end{figure}

\begin{figure}[!htbp]
\centering
% p04_courtyard_bicycle
\includegraphics[width=\linewidth]{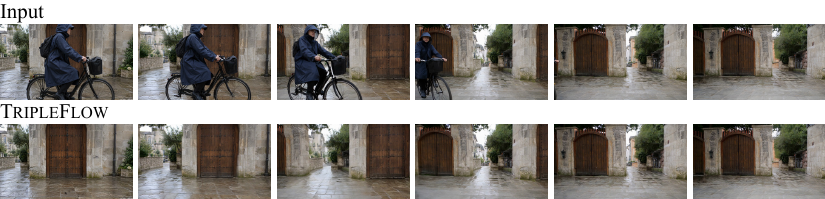}
\caption{Remove the cyclist and bicycle from the courtyard.}
\label{fig:app-p04-courtyard-bicycle}
\end{figure}

\begin{figure}[!htbp]
\centering
% p09_gallery
\includegraphics[width=\linewidth]{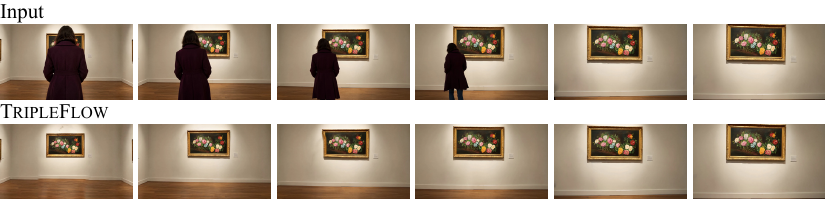}
\caption{Remove the visitor from the art gallery.}
\label{fig:app-p09-gallery}
\end{figure}

\begin{figure}[!htbp]
\centering
% p18_farm_horse
\includegraphics[width=\linewidth]{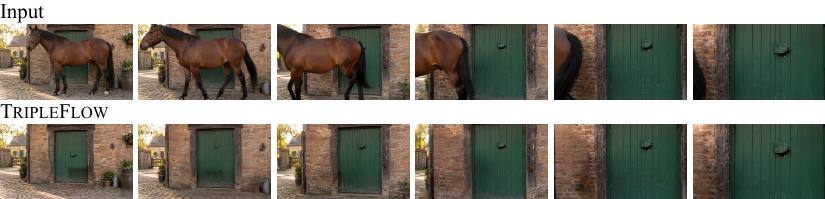}
\caption{Remove the horse from the farmyard.}
\label{fig:app-p18-farm-horse}
\end{figure}

\begin{figure}[!htbp]
\centering
% M_0017
\includegraphics[width=\linewidth]{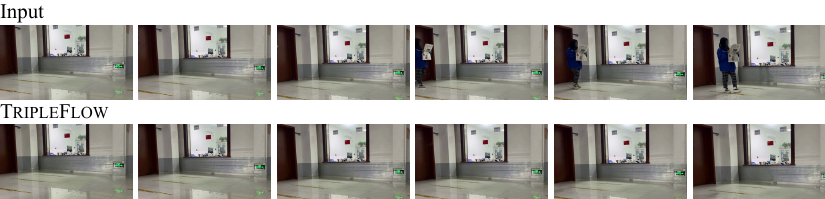}
\caption{Remove the person holding a newspaper.}
\label{fig:app-M-0017}
\end{figure}

\begin{figure}[!htbp]
\centering
% p10_observatory
\includegraphics[width=\linewidth]{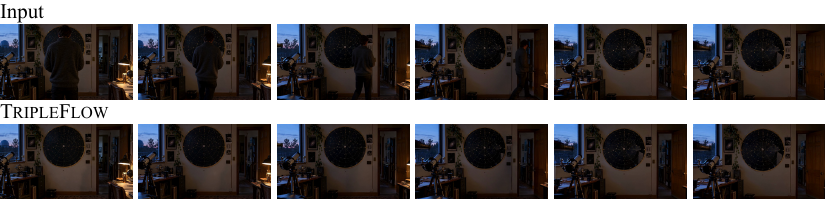}
\caption{Remove the person from the observatory.}
\label{fig:app-p10-observatory}
\end{figure}

\begin{figure}[!htbp]
\centering
% me01_bicycle_plaza
\includegraphics[width=\linewidth]{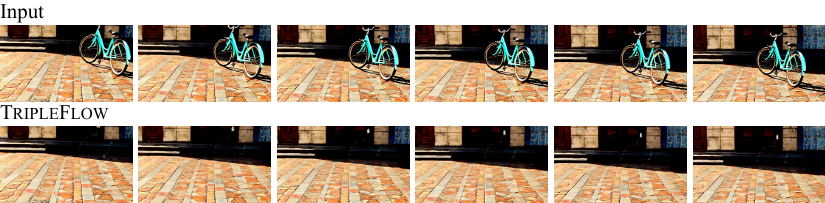}
\caption{Remove the bicycle from the plaza.}
\label{fig:app-me01-bicycle-plaza}
\end{figure}

\clearpage
\section{Additional Qualitative Comparisons}
\label{app:qualitative_comparisons}
We compare five methods on five selected synthetic scenes. Each row
shows six corresponding frames from the same video, including the
first and last frames. The ContextFlow rows use the stored 5B runs
with independently generated MagicQuill first frames; OmniEraser uses
$512\times512$ inputs. These are qualitative examples rather than an
aggregate performance evaluation.

\begin{figure}[!htbp]
\centering
% t06_forklift_shelves
\includegraphics[width=\linewidth]{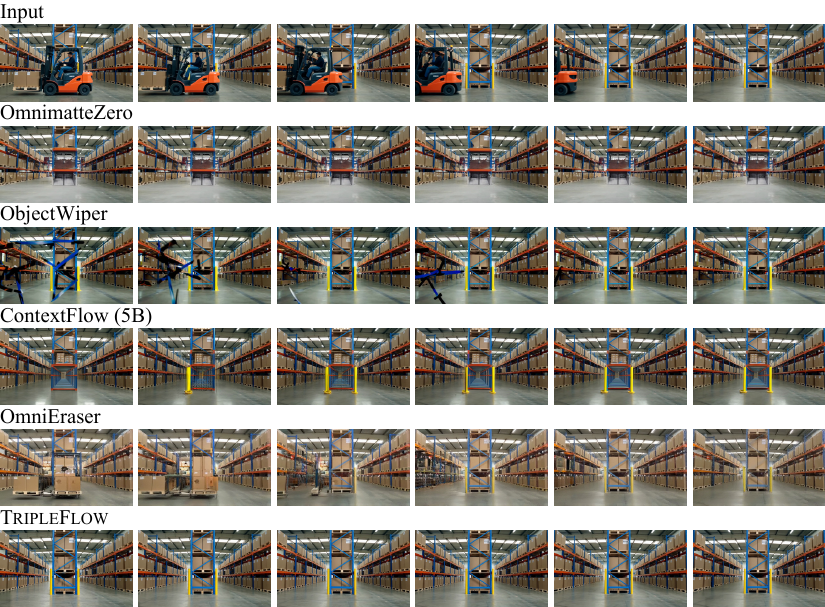}
\caption{Remove the forklift and driver from the warehouse.}
\label{fig:app-t06-forklift-shelves}
\end{figure}

\begin{figure}[!htbp]
\centering
% t02_bookcase
\includegraphics[width=\linewidth]{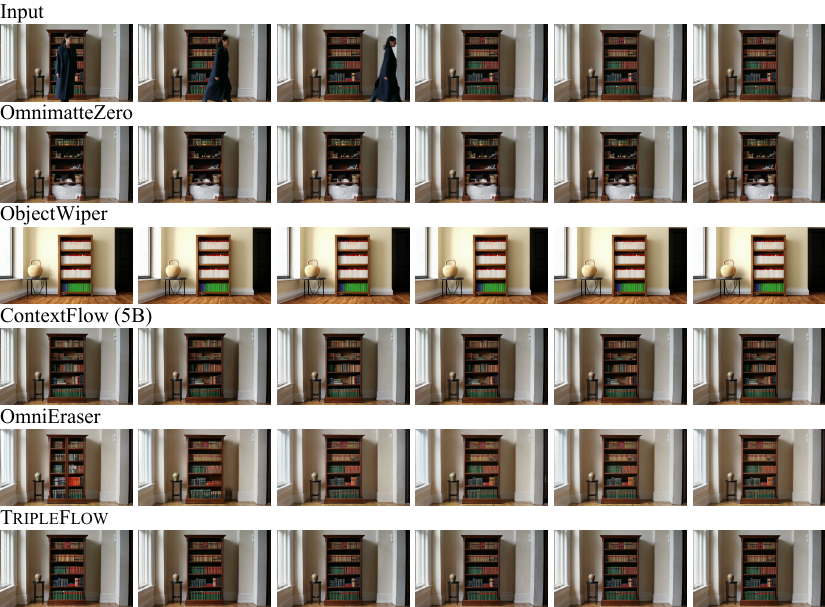}
\caption{Remove the person in front of the bookcase.}
\label{fig:app-t02-bookcase}
\end{figure}

\begin{figure}[!htbp]
\centering
% t01_carved_door
\includegraphics[width=\linewidth]{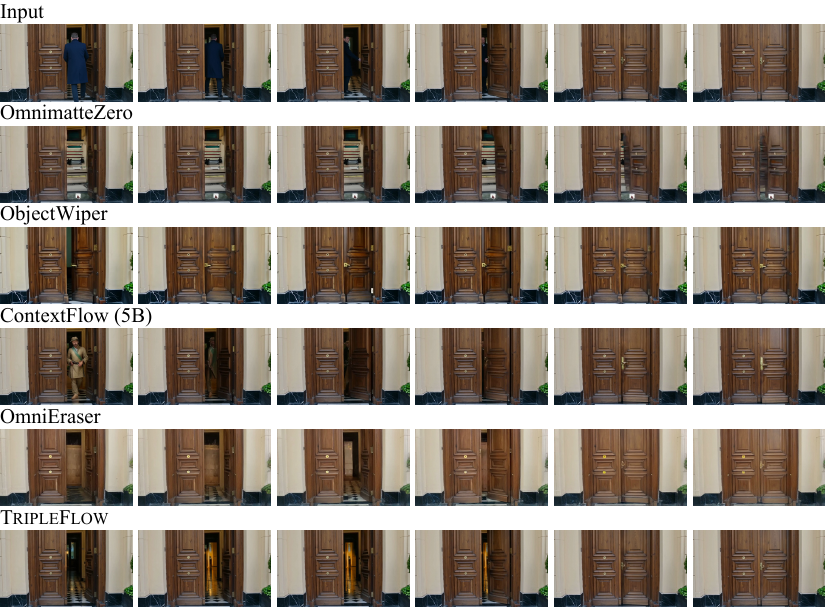}
\caption{Remove the person from the doorway.}
\label{fig:app-t01-carved-door}
\end{figure}

\begin{figure}[!htbp]
\centering
% t09_mirror_entryway
\includegraphics[width=\linewidth]{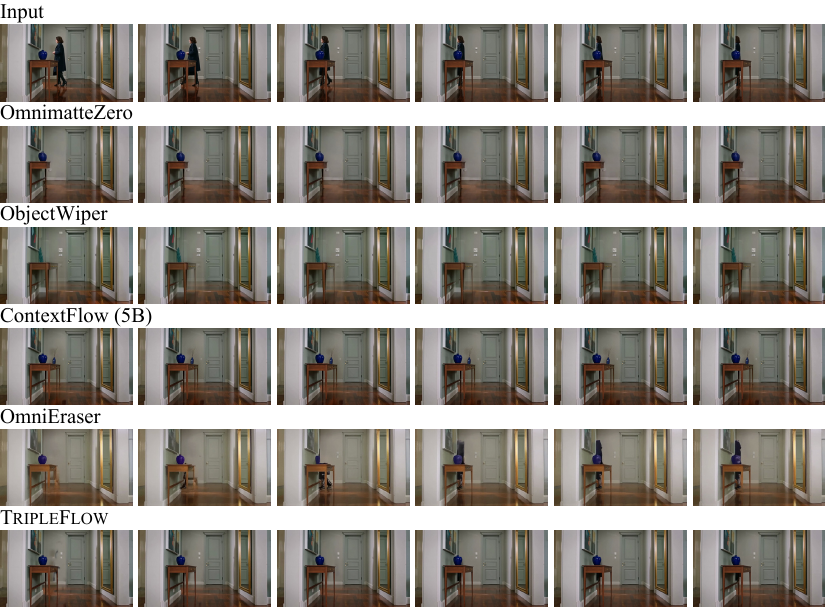}
\caption{Remove the person from the entryway.}
\label{fig:app-t09-mirror-entryway}
\end{figure}

\begin{figure}[!htbp]
\centering
% t04_bench_bicycle
\includegraphics[width=\linewidth]{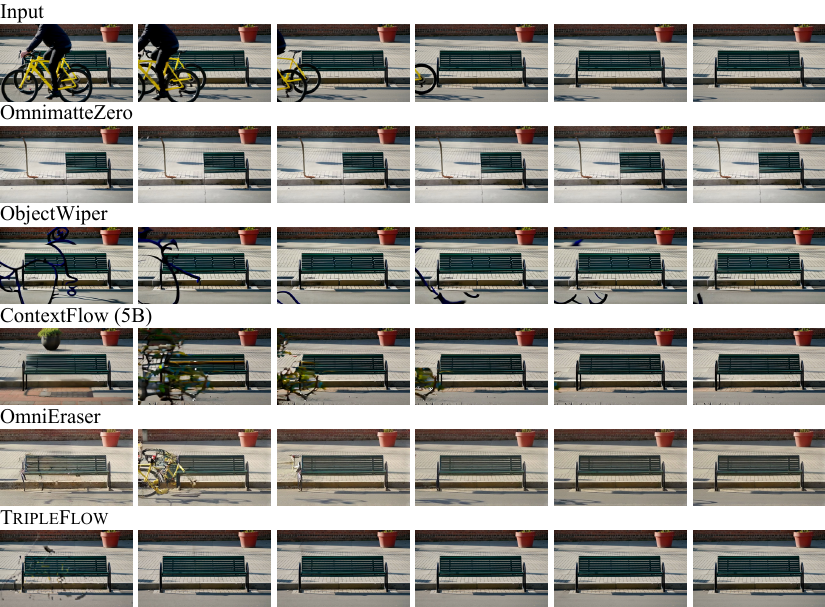}
\caption{Remove the bicycle in front of the bench.}
\label{fig:app-t04-bench-bicycle}
\end{figure}

\clearpage
% END SIX-FRAME QUALITATIVE APPENDIX

\bibliographystyle{iclr2027_conference}
\end{document}